%% file: mmdropout.tex
\pdfoutput=1

\documentclass[11pt]{article}

\usepackage[]{acl}

\usepackage{times}
\usepackage{latexsym}

\usepackage[T1]{fontenc}

\usepackage[utf8]{inputenc}

\usepackage{microtype}

\usepackage{array,multirow}
\usepackage{graphicx}
\usepackage{bm}
\usepackage{amsmath}
\usepackage{amssymb}
\usepackage{amsfonts}
\usepackage{url}
\usepackage{algorithm}
\usepackage[noend]{algpseudocode}
\usepackage {arydshln}

\usepackage{comment}

\usepackage[normalem]{ulem}
\useunder{\uline}{\ul}{}

\renewcommand{\u}{\underline}
\renewcommand{\b}{\textbf}

\title{MaxMatch-Dropout: Subword Regularization for WordPiece}

\author{Tatsuya Hiraoka \\
  Tokyo Institute of Technology\Thanks{ The author is currently affiliated with Fujitsu Limited. This work was carried out at the Tokyo Institute of Technology.} \\
  \texttt{hiraoka.tatsuya@fujitsu.com}
 }

\begin{document}
\maketitle
\begin{abstract}
We present a subword regularization method for WordPiece, which uses a maximum matching algorithm for tokenization.
The proposed method, MaxMatch-Dropout, randomly drops words in a search using the maximum matching algorithm.
It realizes finetuning with subword regularization for popular pretrained language models such as BERT-base.
The experimental results demonstrate that MaxMatch-Dropout improves the performance of text classification and machine translation tasks as well as other subword regularization methods.
Moreover, we provide a comparative analysis of subword regularization methods: subword regularization with SentencePiece (Unigram), BPE-Dropout, and MaxMatch-Dropout.
\end{abstract}

\section{Introduction}

Subword regularization~\cite{kudo2018subword} is a well-known technique for improving the performance of NLP systems, whereby
a model is trained with various tokenizations that are sampled for each training epoch.
This approach provides data augmentation and model robustness against tokenization differences.

\newcite{kudo2018subword} first introduced subword regularization using a unigram language model that was included in their tokenization tool, namely SentencePiece~\cite{kudo2018sentencepiece}, and reported its effectiveness on machine translation tasks.
\newcite{provilkov2019bpe} proposed a subword regularization method for byte pair encoding (BPE) known as BPE-Dropout and demonstrated the superiority of their method over that using the unigram language model in machine translation tasks.
Moreover, subword regularization contributes to the performance improvement of text classification tasks~\cite{hiraoka2019stochastic}.

\begin{figure}[t]
\centering
\includegraphics[scale=0.6]{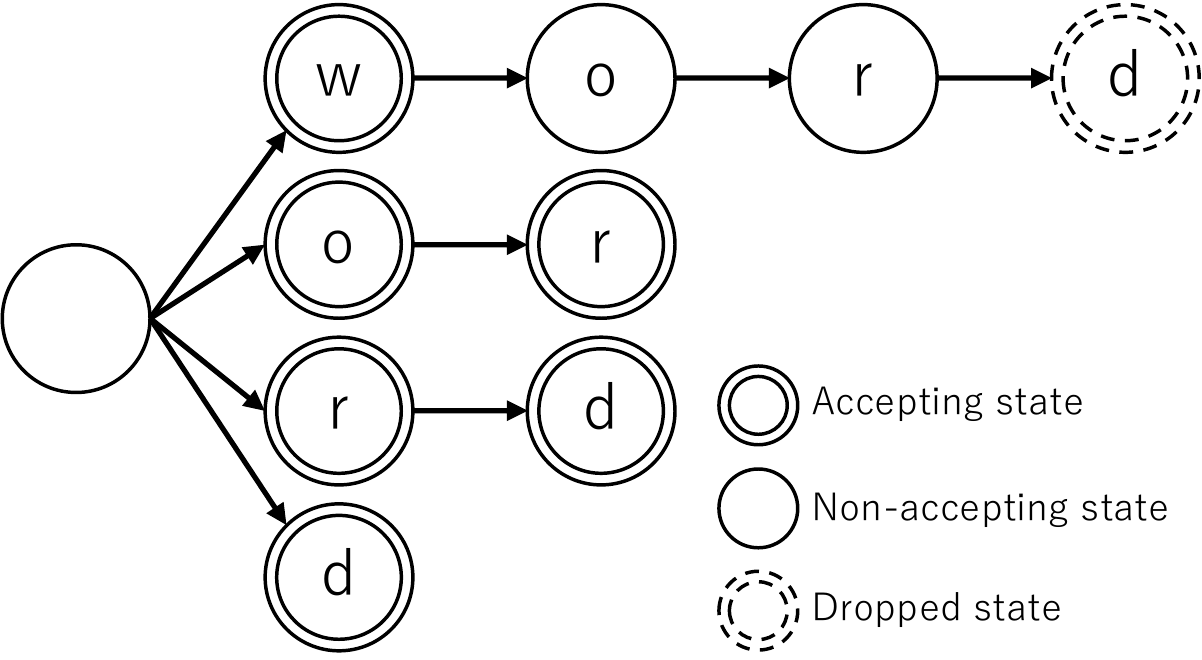}
\caption{
    MaxMatch-Dropout randomly removes accepting states in the trie.
    In this figure, a state corresponding to ``word'' is dropped and a single input ``word'' is tokenized as ``w, or, d.''
}
\label{fgr:outline}
\end{figure}

As subword regularization is implemented as a modification of a tokenizer, each method is specialized to a particular tokenizer type.
For example, the original subword regularization~\cite{kudo2018subword} is specialized to a tokenizer that uses the unigram language model and BPE-Dropout is specialized to the BPE-based tokenizer.
However, these existing subword regularization tools cannot be directly applied to the other common tokenizers such as WordPiece~\cite{song2020linear}.

WordPiece is a tokenizer that is based on the maximum matching algorithm.
It is used as the default tokenizer for the popular pretrained language model BERT~\cite{devlin2018bert}.
Although the widely used BERT models (e.g., BERT-base) can improve the performance of various NLP tasks, subword regularization cannot be used for the finetuning of the model because no subword regularization method exists for WordPiece.
The use of subword regularization for the finetuning of pretrained models with WordPiece may result in a further performance improvement. 

In this paper, we present a simple modification of WordPiece for the use of subword regularization.
The proposed method, which is known as MaxMatch-Dropout, randomly drops words in a vocabulary during the tokenization process.
That is, MaxMatch-Dropout randomly removes accepting states from a trie for tokenization.
The experimental results demonstrate that MaxMatch-Dropout improves the performance of text classification and machine translation in several languages, as well as other subword regularization methods.
Furthermore, MaxMatch-Dropout contributes to a further performance improvement with pretrained BERT on text classification in English, Korean, and Japanese.

\section{Maximum Matching}
\label{sec:maxmatch}
A simple modification to the maximum matching algorithm is implemented so that MaxMatch-Dropout can realize subword regularization.
Prior to explaining the modification, we briefly review the maximum matching on which the proposed method is based\footnote{\newcite{song2020linear} explains the efficient implementation of the maximum matching in detail.}.

Given a vocabulary and a single word, the maximum matching searches the longest subword in the vocabulary and greedily tokenizes the word into a sequence of subwords from beginning to end.
For example, let the vocabulary be composed of \{a, b, c, d, abc, bcd\}. 
The tokenizer with the maximum matching divides a word ``abcd'' into ``abc, d''\footnote{We do not use special tokens for a subword that begins in the middle of a word (e.g., ``\#\#'') for simple explanation.}.
As the maximum matching searches subwords from the beginning of the word, this word is not tokenized as ``a, bcd.''
When an input word includes an unknown character, such as ``abce,'' the tokenizer replaces this word with a special token, ``[UNK].'' 
This tokenization process is usually implemented using a trie.
The detailed tokenization process using the maximum matching for this example with the trie (Figure \ref{fgr:mm}) is described in Appendix \ref{sec:mm_detailed}.

\begin{algorithm}[t]
\small
\caption{Algorithm for Word Tokenization}
\label{alg:mmdropout}
\begin{algorithmic}[1]
\Require Single Word $w$, Vocabulary $V$,  Dropout Rate $q$.
\State $S \leftarrow$ Empty List
\State Index of Characters $i \leftarrow 1$
\While {$i < |w|$}
    \State Subword $s \leftarrow \varnothing$
    \For {$j=1$ to $|w|-i$}
        \If {$w_{i:i+j} \in V$ and $\mathrm{Ber}(1-q)$}
            \State $s \leftarrow w_{i:i+j}$
        \EndIf
    \EndFor
    \If {$s = \varnothing$}
        \Return [UNK]
    \Else
        \State Add $s$ to $S$
        \State $i \leftarrow i + |s|$
    \EndIf
\EndWhile
\Return $S$
\end{algorithmic}
\end{algorithm}

\section{Proposed Method: MaxMatch-Dropout}
\label{sec:proposed}
The proposed method extends the maximum matching with an additional dropout process.
This method randomly replaces accepting states into non-accepting states with \textbf{dropped states}.
That is, accepting tokens are randomly skipped with a specified probability $q$, where $q$ is a hyperparameter.

Figure \ref{fgr:outline} depicts the tokenization process of a word ``word'' with a vocabulary that includes \{w, o, r, d, or, rd, word\}.
Although the maximum matched subword beginning with the first character is ``word'' in the vocabulary, in this case, the state corresponding to ``word'' is dropped.
Thus, the latest accepted subword ``w'' is yielded and the next matching begins from the second character.
Finally, the tokenization process results in ``w, or, d.''

This process is also outlined in Algorithm \ref{alg:mmdropout}~\footnote{Algorithm \ref{alg:mmdropout} does not use a trie for simple explanation.}.
In the algorithm, $w_{i,i+j}$ denotes a subword beginning from the $i$-th character and ending with the $(i+j-1)$-th character in the word $w$, where
$|w|$ and $|s|$ are the lengths of the input word and subword, respectively.
Moreover, $Ber(1-q)$ denotes a Bernoulli distribution that returns $1$ with a probability of $1-q$.

The tokenization process of MaxMatch-Dropout is detailed in Table \ref{tbl:w_or_d} of Appendix \ref{sec:mm_detailed}.
The difference between MaxMatch-Dropout and the original maximum matching can be observed by comparing Tables \ref{tbl:word} and \ref{tbl:w_or_d}.

The regularization strength can be tuned using the hyperparameter $q$.
The proposed method is equivalent to the original maximum matching with $q=0.0$, and it tokenizes a word into characters with $q=1.0$ if all characters are included in the vocabulary.

The official code is available at \url{https://github.com/tatHi/maxmatch_dropout}.

\input{tables/table_tc}

\section{Experiments}
We conducted experiments on text classification and machine translation tasks to validate the performance improvement provided by MaxMatch-Dropout.

We used two tokenizers and subword regularization methods as a reference for both tasks: SentencePiece (Unigram)~\cite{kudo2018sentencepiece} with subword regularization (Sub. Reg.)~\cite{kudo2018subword} and BPE~\cite{sennrich2016neural} with BPE-Dropout~\cite{provilkov2019bpe}.
We employed WordPiece~\cite{song2020linear}, which was implemented by HuggingFace~\cite{wolf-etal-2020-transformers}, as a basic tokenizer for the proposed MaxMatch-Dropout~\footnote{Table \ref{tbl:tokenization} in the Appendix presents tokenization examples for each tokenizer.}.

We set the vocabulary size of each tokenizer to be equal to compare the three methods as fairly as possible.
The vocabulary of each tokenizer included all characters that appeared in the training splits.
We selected the hyperparameters for the subword regularization (e.g., $q$ of MaxMatch-Dropout) according to the performance on the development splits.
Note that we could not fairly compare the performance of MaxMatch-Dropout to that of other subword regularization methods because they are based on different tokenizers and vocabularies.
WordPiece was used as the baseline for MaxMatch-Dropout to investigate whether the method could successfully perform subword regularization and improve the performance similarly to other methods.

\subsection{Text Classification}
\paragraph{Datasets}
We exploited text classification datasets in three languages: English, Korean, and Japanese.
\textbf{APG} and \textbf{APR} are genre prediction and rating prediction, respectively, on review texts that were created from the Amazon Product Dataset~\cite{he2016ups}.
\textbf{TS} is a sentiment classification for tweets~\footnote{\url{https://www.kaggle.com/c/twitter-sentiment-analysis2}}.
We also employed \textbf{QNLI}~\cite{rajpurkar2016squad}, \textbf{QQP}~\cite{chen2018quora}, \textbf{RTE}~\cite{bentivogli2009fifth}, and \textbf{SST-2}~\cite{socher2013recursive} from the GLUE benchmark~\cite{wang2018glue}.
\textbf{NLI}, \textbf{STS}, and \textbf{YNAT} are text classification datasets that are included in Korean GLUE (KLUE)~\cite{park2021klue}.
\textbf{TR}~\cite{suzuki2019filtering} and \textbf{WRIME}~\cite{kajiwara2021wrime} are sentiment classification datasets for tweets in Japanese.
We used the original development sets as test sets and exploited a randomly selected 10\% of the original training sets as development sets for the datasets in GLUE and KLUE owing to the numerous experimental trials.

\paragraph{Setup}
We used two backbones for the text classification: BiLSTM~\cite{hochreiter1997long,graves2005framewise} and BERT~\cite{devlin2018bert}.
We employed BERT-base-cased\footnote{\url{https://huggingface.co/bert-base-cased}}, BERT-kor-base\footnote{\url{https://huggingface.co/kykim/bert-kor-base}}\cite{kim2020lmkor}, and BERT-base-Japanese-v2\footnote{\url{https://huggingface.co/cl-tohoku/bert-base-japanese-v2}} for the English, Korean, and Japanese datasets, respectively.
All of these BERT models employ WordPiece as their tokenizers, and we finetuned them using MaxMatch-Dropout.
We set the maximum number of training epochs to 20 for BiLSTM and the finetuning epochs to 5 for BERT.
The trained model with the highest score in the development split was selected and evaluated on the test split.
We selected the vocabulary sizes according to the performance on the development splits when using WordPiece without MaxMatch-Dropout.
The selected vocabulary sizes were applied to all tokenizers.

\paragraph{Results} 
Table \ref{tbl:tc} presents the experimental results for the text classification.
The table demonstrates that MaxMatch-Dropout (MM-Dropout) improved the performance as well as the other subword regularization methods.
In addition to the improvement in the BiLSTM-based classifiers, MaxMatch-Dropout enhanced the performance of the BERT-based classifiers.
These results indicate that MaxMatch-Dropout is a useful subword regularization method for WordPiece as well as effective for BERT.

\input{tables/table_mt}

\subsection{Machine Translation}
\paragraph{Datasets} 
We employed three language pairs for the machine translation tasks: the De-En, Vi-En, and Zh-En pairs from the IWSLT corpora.
We selected these datasets because subword regularization is particularly efficient in low-resource environments~\cite{kudo2018subword,hiraoka2021joint,takase2022single}.

\paragraph{Setup} 
We applied the Transformer~\cite{vaswani2017attention}, which was implemented by Fairseq~\cite{ott2019fairseq}, for the IWSLT settings.
We trained the model with 100 epochs and averaged the parameters of the final 10 epochs.
We evaluated the performance on the Chinese dataset using character-level BLEU.
Following \newcite{provilkov2019bpe}, we set the vocabulary size to 4K for English, German, and Vietnamese, and 16K for Chinese.

\paragraph{Results} 
Table \ref{tbl:mt} displays the experimental results for the machine translation.
The table demonstrates that MaxMatch-Dropout improved the performance in all language pairs.
The results indicate that the proposed method is effective for machine translation as well as existing subword regularization methods.

\section{Discussion}
\subsection{Effect of Hyperparameters}
\label{sec:hyp}
\begin{figure}[t]
\centering
\includegraphics[scale=0.66]{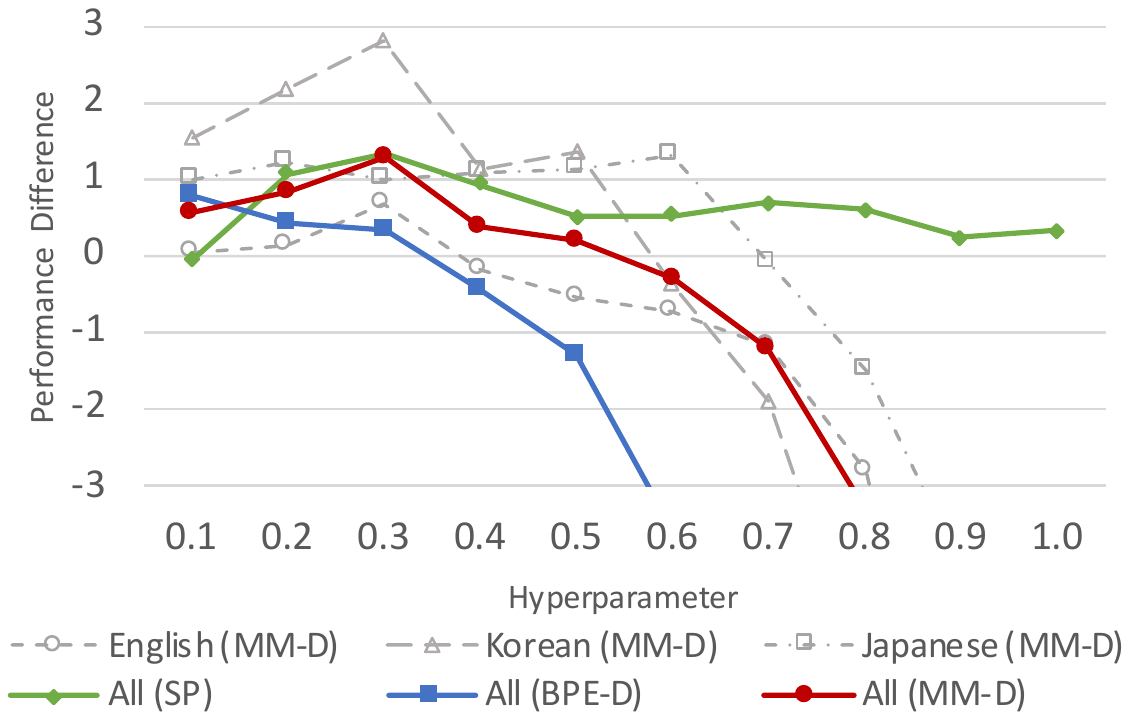}
\caption{
    Performance differences with and without subword regularization against hyperparameters and for different languages on text classification datasets.
    MM-D, SP, and BPE-D denote MaxMatch-Dropout, SentencePiece (Unigram), and BPE-Dropout, respectively.
}
\label{fgr:hyp_tc}
\end{figure}

Figure \ref{fgr:hyp_tc} depicts the averaged performance improvement over several text classification datasets against different hyperparameters.
The figure indicates that the subword regularization of SentencePiece (Unigram) was the most robust against the hyperparameters among the three methods.
Although both BPE-Dropout and MaxMatch-Dropout could realize subword regularization using the dropout technique for the tokenization strategy,
MaxMatch-Dropout was more robust against the hyperparameters than BPE-Dropout.
This result demonstrates that a performance improvement can be achieved in WordPiece-based systems using MaxMatch-Dropout with approximately selected hyperparameters (e.g., $q<0.5$).

Figure \ref{fgr:hyp_tc} also shows the averaged performance on the datasets in each language against the hyperparameters of MaxMatch-Dropout (dashed lines).
It can be observed that MaxMatch-Dropout was more effective for Asian languages than English.
It is considered that this is because Korean and Japanese contain various types of n-grams and many tokenization candidates exist for a single sentence compared to English.

\subsection{Token Length}
In this subsection, we analyze the token length in the sampled tokenizations.
We sampled the tokenization of the training dataset (APG) with three subword regularization methods and counted the token lengths for 10 trials.

\begin{figure}[t]
\centering
\includegraphics[scale=0.73]{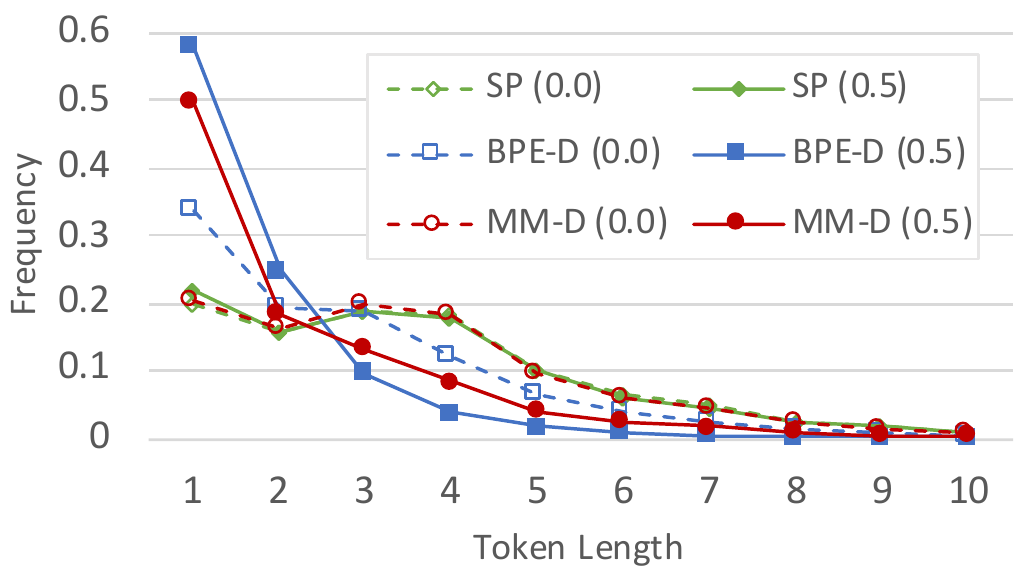}
\caption{
    Frequency of token lengths with each subword regularization method on APG dataset (English).
    0.0 denotes the vanilla settings without subword regularization.
    0.5 indicates subword regularization when the hyperparameter was 0.5 (e.g., $q=0.5$).
    MM-D, SP, and BPE-D denote MaxMatch-Dropout, SentencePiece (Unigram), and BPE-Dropout, respectively.
}
\label{fgr:length_tc}
\end{figure}

Figure \ref{fgr:length_tc} presents the frequency of token lengths in the tokenized training datasets with/without subword regularization.
The figure indicates that the length frequency did not change, regardless of the use of subword regularization, when SentencePiece (Unigram) was applied.
In contrast, both MaxMatch-Dropout (MM-D) and BPE-Dropout (BPE-D) yielded many characters when the hyperparameter was 0.5, because they are based on the token-level dropout and yield characters when the hyperparameter is 1.0.
However, the frequency curve of MaxMatch-Dropout was gentler than that of BPE-Dropout.
We believe that this tendency aided in the robustness of the MaxMatch-Dropout performance, as reported in Section \ref{sec:hyp}.

\section{Conclusion}

We have introduced a subword regularization method for WordPiece, which is a common tokenizer for BERT.
The proposed method, MaxMatch-Dropout, modifies the tokenization process using the maximum matching to drop words in the vocabulary randomly.
This simple modification can realize subword regularization for WordPiece.
Furthermore, the experimental results demonstrated that MaxMatch-Dropout can improve the performance of BERT.
MaxMatch-Dropout is also effective in the training of text classification tasks without BERT and machine translation tasks, as well as existing subword regularization methods.

\section*{Acknowledgement}
This work was supported by JST, ACT-X Grant Number JPMJAX21AM, Japan.

\bibliography{mmdropout}
\bibliographystyle{acl_natbib}

\clearpage
\appendix

\section{Maximum Matching in Detail}
\label{sec:mm_detailed}
\begin{figure}[t]
\centering
\includegraphics[scale=0.65]{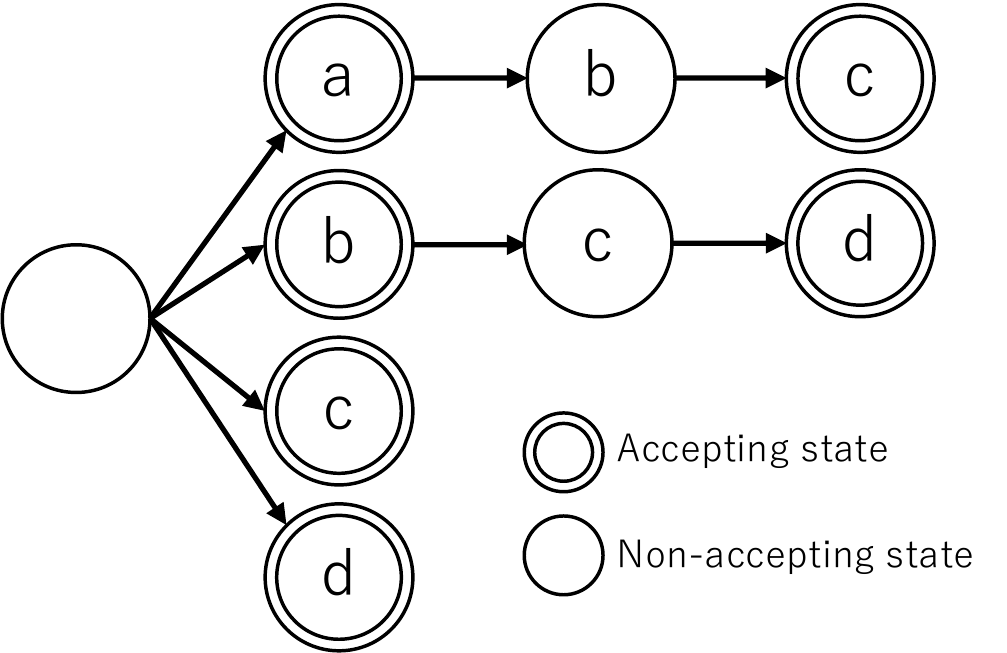}
\caption{
    Trie for vocabulary including tokens \{a, b, c, d, abc, bcd\}.
}
\label{fgr:mm}
\end{figure}

As described in Section \ref{sec:maxmatch}, a trie is generally used to tokenize an input word with the maximum matching algorithm.
Figure \ref{fgr:mm} depicts the trie corresponding to the vocabulary that includes six tokens: \{a, b, c, d, abc, bcd\}.
The tokenization process using this trie for the input words ``abcd'' and ``abce'' is presented in Tables \ref{tbl:abcd} and \ref{tbl:abce}, respectively.

Table \ref{tbl:w_or_d} details the operation for tokenizing an input word ``word'' into ``w, or, d'' using the proposed MaxMatch-Dropout, as outlined in Section \ref{sec:proposed}.
Table \ref{tbl:word} describes the tokenization process using the original maximum matching for Figure \ref{fgr:outline} without the dropout process.
Therefore, the difference in the tokenization process between the original maximum matching and MaxMatch-Dropout can be observed by comparing Tables \ref{tbl:word} and \ref{tbl:w_or_d}.

\input{tables/table_transitions}

\section{Related Work}
This work is related to tokenization methods, which split raw texts into a sequence of tokens.
Three well-known tokenization methods have been employed in recent NLP systems: SentencePiece (Unigram)~\cite{kudo2018sentencepiece}, BPE~\cite{sennrich2016neural}, and WordPiece~\cite{song2020linear}.
SentencePiece (Unigram) is a unigram language model-based tokenizer, whereas BPE employs a frequency-based tokenization technique.
Although both methods are used extensively in many NLP systems, \newcite{bostrom2020byte} reported that the unigram language model-based tokenizer (i.e., SentencePiece (Unigram)) is superior to BPE in several downstream tasks.
Our experimental results in Tables \ref{tbl:tc} and \ref{tbl:mt} also support this finding.

WordPiece\footnote{Although the original term ``wordpiece'' indicates BPE-based tokenization, in this paper, ``WordPiece'' indicates a tokenizer with the maximum matching for BERT.} is another famous tokenizer that is mainly employed by large pretrained models such as BERT~\cite{devlin2018bert}.
As WordPiece is based on the maximum matching algorithm, it is superior to other tokenization methods in terms of the tokenization speed.
In fact, WordPiece is employed in real NLP systems such as Google searching~\cite{song2020linear}. 
However, the experimental results in this study (Table \ref{tbl:tc} and \ref{tbl:mt}) demonstrated that WordPiece is inferior to SentencePiece (Unigram) and BPE in terms of performance.
The proposed method can compensate for this shortcoming without decreasing the inference speed.

\newcite{kudo2018subword} introduced a subword regularization technique for SentencePiece (Unigram) using dynamic programming.
\newcite{provilkov2019bpe} proposed a subword regularization method for BPE using the dropout technique.
This study has introduced a subword regularization method for WordPiece, and presented an in-depth investigation of the three methods in text classification and machine translation.

\section{Contributions}
This study contributes to the NLP community in terms of the following two main points:
\begin{itemize}
    \item A subword regularization method for WordPiece is proposed, which improves the text classification and machine translation performance.
    \item An intensive performance investigation of the three famous tokenization and subword regularization methods used in NLP (i.e., SentencePiece (Unigram), BPE, and WordPiece with subword regularization) is presented.
\end{itemize}

\section{Dataset Statistics}
Table \ref{tbl:data} displays the detailed information of the datasets.
We report the numbers of samples in the training, development, and test splits.
Furthermore, we present the number of label types for text classification datasets.

\input{tables/table_data}

\section{Detailed Experimental Settings}
Tables \ref{tbl:tc_settings} and \ref{tbl:mt_settings} present the detailed settings of the backbone models that were used in text classification and machine translation tasks, respectively.
We used the default values of PyTorch for the hyperparameters that are not described in these tables.
We set the number of tokenization candidates to $\infty$ for the subword regularization of SentencePiece (Unigram).

We selected the hyperparameters for the subword regularization methods (the smoothing parameter for SentencePiece (Unigram) and the dropout probabilities for BPE-Dropout and MaxMatch-Dropout) according to the performance on the development splits in the experiments.
Tables \ref{tbl:tc_hyps} and \ref{tbl:mt_hyps} summarize the selected values of the hyperparameters for the text classification and machine translation, respectively.
Note that the other methods without subword regularization (Unigram, BPE, and WordPiece) do not require these hyperparameters.

\input{tables/table_tc_settings}
\input{tables/table_mt_settings}

\input{tables/table_tc_hyps}
\input{tables/table_mt_hyps}

\input{tables/table_examples}

\end{document}

%% file: tables/table_tc.tex
\begin{table*}[ht]
\small
\centering
\begin{tabular}{l|wc{6.5mm}wc{6.5mm}wc{6.5mm}wc{6.5mm}wc{6.5mm}wc{6.5mm}wc{6.5mm}wc{6.5mm}wc{6.5mm}wc{6.5mm}wc{6.5mm}wc{6.5mm}}
\hline
         & \multicolumn{7}{l}{English}           & \multicolumn{3}{l}{Korean}                 & \multicolumn{2}{l}{Japanese}      \\
         & APG & APR & TS & QNLI & QQP & RTE & SST-2 & NLI & STS & YNAT & TR & WRIME \\
$|V|$    & 32K & 32K & 32K& 32K  & 32K & 12K & 8K    & 24K & 16K & 32K  & 16K& 12K \\
Metric   & F1  & F1  & F1 & Acc. & F1  & Acc.& Acc.  & Acc.& F1  & F1   & F1 & F1    \\
\hline
\textit{BiLSTM} &  &  &  &  &  &  &  &  &  &  &  &  \\
Unigram            & 69.05       &65.85        &76.21        & 66.48   & 83.61       & 49.10   & 80.05       & 41.93   & 67.02     & 68.57       & 86.6 & 46.36 \\
+ Sub. Reg. &\b{\u{70.65}}&\b{\u{66.80}}&\b{\u{77.49}}&\b{66.56}&\b{\u{83.91}}&\b{53.31}&\b{\u{83.30}}&\b{42.84}&\b{68.08}  &\b{\u{73.67}}&\b{87.11}    &\b{\u{49.47}} \\
\hdashline
BPE                      & 67.10       & 64.67       & 75.24       &\b{67.11}&\b{82.82}    & 53.07   & 78.10       & 41.22   &\b{67.42}  & 64.27       & 84.95 & 44.34 \\
+ BPE-Dropout            &\b{\u{68.45}}&\b{\u{65.38}}&\b{\u{76.04}}& 66.69   & 82.69       &\b{53.97}&\b{\u{82.00}}&\b{41.52}& 66.26     &\b{\u{69.12}}&\b{85.68}    &\b{46.01} \\
\hdashline
WordPiece                & 63.17       & 62.97       & 73.14       & 64.04   & 82.11       & 53.55   & 81.04       & 39.96   & 61.75     & 62.44       & 84.95 & 46.36 \\
+ MM-Dropout             &\b{\u{64.90}}&\b{\u{64.36}}&\b{\u{75.22}}&\b{64.28}&\b{82.14}    &\b{53.91}&\b{\u{83.75}}&\b{40.61}&\b{\u{62.88}}&\b{\u{70.08}}&\b{\u{86.98}}&\b{47.28} \\
\hline
\textit{BERT} &  &  &  &  &  &  &  &  &  &  &  &  \\
WordPiece                & 77.28 & 70.99 & 81.93 & 89.45 & 89.83 & 62.00 & 90.97 & 82.18 & 83.22 & 83.96 & 89.08 & 89.08 \\
+ MM-Dropout             & \u{\b{78.55}} & \u{\b{71.68}} & \b{82.08} & \b{89.74} & \textbf{89.86} & \textbf{62.27} & \textbf{91.07} & \textbf{82.19} & \u{\textbf{85.43}} & \u{\textbf{84.31}} & \textbf{89.14} & \textbf{89.14} \\
\hline
\end{tabular}
\caption{
    \label{tbl:tc}
    Experimental results of text classification (averaged scores of five runs).
    The higher scores for the tokenizations with/without subword regularization are indicated in bold.
    The scores that significantly surpassed the results without subword regularization ($p < 0.05$, McNemar’s test) are underlined.
}
\end{table*}

%% file: tables/table_mt.tex
\begin{table}[t]
\small
\centering
\begin{tabular}{wl{18.5mm}|wc{4.8mm}wc{4.8mm}wc{4.8mm}wc{4.8mm}wc{4.8mm}wc{4.8mm}}
\hline
              & \multicolumn{2}{l}{IWSLT14}           & \multicolumn{4}{l}{IWSLT15} \\
              & DeEn    & EnDe  & ViEn  & EnVi  & ZhEn  & EnZh  \\
\hline
Unigram      & 36.55 & 27.89 & 30.28 & 29.39 & 22.64 & 20.55 \\
+ Sub. Reg.  & \u{\b{38.50}} & \u{\b{29.45}} & \u{\b{31.58}} & \u{\b{30.96}} & \u{\b{23.81}} & \u{\b{21.79}} \\
\hdashline
BPE & 35.77 & 27.87 & 30.05 & 29.25 & 18.80 & 20.61 \\
+ BPE-Dropout & \u{\b{37.81}} & \u{\b{29.15}} & \u{\b{31.39}} & \u{\b{31.23}} & \u{\b{20.67}} & \u{\b{22.02}} \\
\hdashline
WordPiece & 36.22 & 27.58 & 30.13 & 29.40 & 17.24 & 20.45 \\
+ MM-Dropout & \u{\b{38.30}} & \u{\b{29.54}} & \u{\b{31.71}} & \u{\b{31.14}} & \u{\b{18.21}} & \u{\b{21.55}} \\
\hline
\end{tabular}
\caption{
    \label{tbl:mt}
    Experimental results of machine translation (averaged scores of three runs). ScareBLEU~\cite{post2018call} was used as the metric.
    Scores that significantly surpassed the results without subword regularization ($p < 0.05$, bootstrap resampling \cite{koehn2007moses}) are underlined.
}
\end{table}

%% file: tables/table_transitions.tex
\begin{table}[t]
\centering
\small
\begin{tabular}{c|l|l}
\hline
Read & Action                          & Output \\ \hline
a    & Accept "a"                      &        \\
b    & Non-accept "ab"                 &        \\
c    & Accept "abc"                    &        \\
d    & Reject the transition to "abcd" &     \\
     & \& Yield the latest subword     & abc       \\
d    & Accept "d"                      &        \\
\$   & Reject the transition to "d\$"   &  \\
     & \& Yield the latest subword     & abc, d       \\ \hline
\end{tabular}
\caption{
    Operation for tokenizing input word ``abcd'' into ``abc, d'' using trie shown in Figure \ref{fgr:mm}. ``\$'' denotes a special symbol indicating the end of the word.
}
\label{tbl:abcd}
\end{table}

\begin{table}[t]
\centering
\small
\begin{tabular}{c|l|l}
\hline
Read & Action                          & Output \\ \hline
a    & Accept "a"                      &        \\
b    & Non-accept "ab"                 &        \\
c    & Accept "abc"                    &        \\
e    & Reject the transition to "abcd" &     \\
     & \& Yield the latest subword     & abc       \\
e    & Detect an OOV character            &        \\
     & \& Output [UNK]                & [UNK]       \\ \hline
\end{tabular}
\caption{
    Operation for tokenizing input word ``abce'' including out-of-vocabulary (OOV) character into ``[UNK]'' using trie shown in Figure \ref{fgr:mm}.
}
\label{tbl:abce}
\end{table}

\begin{table}[ht]
\centering
\small
\begin{tabular}{c|l|l}
\hline
Read & Action                            & Output \\ \hline
w    & Accept "w"                        &        \\
o    & Non-accept "wo"                   &        \\
r    & Non-accept "wor"                  &        \\
d    & Accept "word"                     &        \\
\$   & Reject the transition to "word\$" &        \\
     & \& Yield the latest subword       & word   \\ \hline
\end{tabular}
\caption{
    Operation for tokenizing input word ``word'' by applying original maximum matching (i.e., the operation without any dropout process) for trie shown in Figure \ref{fgr:outline}.
    ``\$'' denotes a special symbol indicating the end of the word.
}
\label{tbl:word}
\end{table}

\begin{table}[ht]
\centering
\small
\begin{tabular}{c|l|l}
\hline
Read & Action                                & Output   \\ \hline
w    & Accept "w"                            &          \\
o    & Non-accept "wo"                       &          \\
r    & Non-accept "wor"                      &          \\
d    & \textbf{(Randomly) Non-accept "word"} &          \\
\$   & Reject the transition to "word\$"     &          \\
     & \& Yield the latest subword           & w        \\
o    & Accept "o"                            &          \\
r    & Accept "or"                           &          \\
d    & Reject the transition to "ord"        &          \\
     & \& Yield the latest subword           & w, or    \\
d    & Accept "d"                            &          \\
\$   & Reject the transition to "d\$"        &          \\
     & \& Yield the latest subword           & w, or, d \\ \hline
\end{tabular}
\caption{
    Operation for tokenizing input word ``word'' using trie for MaxMatch-Dropout shown in Figure \ref{fgr:outline}.
    ``\$'' denotes a special symbol indicating the end of the word.
}
\label{tbl:w_or_d}
\end{table}

%% file: tables/table_data.tex
\begin{table}[h]
\small
\centering
\begin{tabular}{lrrrr}
\hline
Dataset & Train & Dev. & Test & Labels \\
\hline
\multicolumn{5}{l}{\textit{English Text Classification}} \\
APG     & 96,000&12,000&12,000& 24 \\
APR     & 96,000&12,000&12,000& 5 \\
TS      & 80,000&10,000&10,000& 2       \\
QNLI    &188,536&10,475&5,463& 2   \\
QQP     &327,461&36,385&40,430& 2       \\
RTE     &2,241&249&277& 2       \\
SST-2   &60,614&6,735&872& 2       \\
\hdashline
\multicolumn{5}{l}{\textit{Korean Text Classification}} \\
NLI     &22,498&2,500&3,000& 3       \\
STS     &10,501&1,167&519& 2       \\
YNAT    &41,110&4,568&9,107& 7       \\
\hdashline
\multicolumn{5}{l}{\textit{Japanese Text Classification}} \\
TR      &129,747&16,218&16,219& 3       \\
WRIME   &30,000&2,500&2,500& 5      \\
\hdashline
\multicolumn{5}{l}{\textit{Machine Translation}} \\
DeEn    &160,239&7,283 &6,750 & -      \\
ViEn    &130,933&768.  &1,268 & -      \\
ZhEn    &209,941&887.  &1,261 & -      \\
\hline
\end{tabular}
\caption{
    \label{tbl:data}
    Statistics of datasets.
}
\end{table}

%% file: tables/table_tc_settings.tex
\begin{table}[t]
\centering
\small
\begin{tabular}{c|rr}
\hline
Parameter                & BiLSTM & BERT    \\ \hline
Embedding Size           & 64     & 768     \\
BiLSTM/BERT Hiden Size   & 256    & 768     \\
\# of BiLSTM/BERT Layers & 1      & 12      \\
Dropout Rate             & 0.5    & 0.1     \\
Optimizer                & Adam   & AdamW   \\
Learning Rate            & 0.001  & 0.00002 \\
\hline
\end{tabular}
\caption{
    Overview of hyperparameters for backbone models of text classification tasks.
}
\label{tbl:tc_settings}
\end{table}

%% file: tables/table_mt_settings.tex
\begin{table}[h]
\centering
\small
\begin{tabular}{l|r}
\hline
Parameter                         & Transformer               \\ \hline
Enc/Dec Embedding Size            & 512                 \\
Enc/Dec FFN Embedding Size        & 1,024               \\
\# of Enc/Dec Attention Heads & 4                   \\
\# of Enc/Dec Layers          & 6                   \\
Clipping Norm                     & 0.0                 \\
Dropout Rate                      & 0.3                 \\
Weight Decay                      & 0.0001              \\
Max Tokens for Mini-Batch         & 1,000               \\
Optimizer                         & Adam                \\
$\beta_1$ and $\beta_2$ for Adam  & 0.9, 0.98           \\
Learning Rate                     & 0.0005              \\
Learning Rate Scheduler           & Inverse Square Root \\
Warming-Up Updates                & 4,000               \\ \hline
\end{tabular}
\caption{
    Overview of hyperparameters for backbone model of machine translation tasks.
}
\label{tbl:mt_settings}
\end{table}

%% file: tables/table_tc_hyps.tex
\begin{table*}[b]
\small
\centering
\begin{tabular}{l|rrrrrrrrrrrr}
\hline
                  & \multicolumn{7}{l}{English}                & \multicolumn{3}{l}{Korean} & \multicolumn{2}{l}{Japanese} \\
                  & APG & APR & TS  & QNLI & QQP & RTE & SST-2 & NLI     & STS    & YNAT    & TR           & WRIME         \\ \hline
\textit{BiLSTM}            &     &     &     &      &     &     &       &         &        &         &              &               \\
Unigram+Sub. Reg. & 0.2 & 0.2 & 0.2 & 0.6  & 0.9 & 0.3 & 0.2   & 0.9     & 0.3    & 0.3     & 0.4          & 1.0           \\
BPE-dropout       & 0.2 & 0.2 & 0.4 & 0.1  & 0.1 & 0.1 & 0.3   & 0.3     & 0.2    & 0.3     & 0.5          & 0.2           \\
MaxMatch-dropout        & 0.2 & 0.3 & 0.6 & 0.1  & 0.1 & 0.3 & 0.4   & 0.4     & 0.2    & 0.3     & 0.4          & 0.6           \\ \hline
\textit{BERT}              &     &     &     &      &     &     &       &         &        &         &              &               \\
MaxMatch-Dropout        & 0.6 & 0.4 & 0.2 & 0.1  & 0.1 & 0.1 & 0.3   & 0.5     & 0.4    & 0.5     & 0.4          & 0.5           \\ \hline
\end{tabular}
\caption{
    \label{tbl:tc_hyps}
    Selected hyperparameters for subword regularization methods in text classification tasks.
}
\end{table*}

%% file: tables/table_mt_hyps.tex
\begin{table*}[b]
\small
\centering
\begin{tabular}{l|rrrrrr}
\hline
                    & \multicolumn{2}{l}{IWSLT14} & \multicolumn{4}{l}{IWSLT15} \\
                    & DeEn         & EnDe         & ViEn  & EnVi  & ZhEn & EnZh \\ \hline
Unigram + Sub. Reg. & 0.3          & 0.3          & 0.4   & 0.3   & 0.2  & 0.2  \\
BPE-Dropout         & 0.1          & 0.2          & 0.2   & 0.2   & 0.3  & 0.2  \\
MaxMatch-Dropout          & 0.3          & 0.3          & 0.4   & 0.1   & 0.1  & 0.2  \\ \hline
\end{tabular}
\caption{
    \label{tbl:mt_hyps}
    Selected hyperparameters for subword regularization methods in machine translation tasks.
    The selected hyperparameters were used for the subword regularization of both the source and target languages.
}
\end{table*}

%% file: tables/table_examples.tex
\begin{table*}[]
\centering
\small
\begin{tabular}{lllll}
\hline
Hyperparameter     & Trial & Unigram+Sub. Reg.                    & BPE-Dropout                                          & MaxMatch-Dropout                                   \\ \hline
No regularization & -     & characteristics                      & characteristics                                      & characteristics                              \\\hdashline
0.1                & 1     & \textbf{character\_i\_s\_t\_ic\_s}   & characteristics                                      & \textbf{characteristic\_s}                            \\
                   & 2     & \textbf{character\_i\_s\_t\_ics}     & characteristics                                      & characteristics                              \\
                   & 3     & \textbf{characteristic\_s}           & characteristics                                      & characteristics                              \\
                   & 4     & \textbf{cha\_rac\_t\_e\_r\_istic\_s} & characteristics                                      & characteristics                              \\
                   & 5     & \textbf{ch\_ar\_act\_e\_r\_istic\_s} & characteristics                                      & characteristics                              \\\hdashline
0.5                & 1     & characteristics                      & characteristics                                      & \textbf{characteristic\_s}                   \\
                   & 2     & characteristics                      & \textbf{c\_har\_ac\_ter\_istics}                     & characteristics                              \\
                   & 3     & characteristics                      & characteristics                                      & \textbf{char\_acter\_istics}                 \\
                   & 4     & characteristics                      & \textbf{char\_ac\_ter\_istics}                       & characteristics                              \\
                   & 5     & \textbf{characteristic\_s}           & \textbf{character\_ist\_ics}                         & characteristics                              \\\hdashline
0.9                & 1     & characteristics                      & \textbf{c\_h\_a\_r\_a\_c\_t\_er\_i\_s\_t\_i\_c\_s} & \textbf{char\_a\_c\_t\_e\_ri\_s\_t\_i\_c\_s} \\
                   & 2     & characteristics                      & \textbf{char\_ac\_t\_er\_ist\_ics}                   & \textbf{c\_har\_a\_c\_t\_e\_r\_istics}       \\
                   & 3     & characteristics                      & \textbf{c\_h\_ar\_a\_c\_t\_er\_i\_s\_t\_ic\_s}       & \textbf{ch\_a\_r\_acter\_i\_s\_t\_i\_c\_s}   \\
                   & 4     & characteristics                      & \textbf{c\_h\_a\_r\_ac\_t\_e\_r\_i\_s\_ti\_c\_s}   & \textbf{character\_i\_s\_t\_i\_cs}           \\
                   & 5     & characteristics                      & \textbf{c\_ha\_ra\_ct\_er\_i\_st\_i\_c\_s}         & \textbf{character\_i\_stic\_s}               \\ \hline
\end{tabular}
\caption{
    Examples of tokenized words using three methods with different hyperparameters for five trials.
    ``\_'' indicates token boundaries.
    The vocabularies for each method were constructed using the APG dataset.
    Sampled tokenizations that differed from the original tokenizations without subword regularization are indicated in bold.
    We removed special symbols indicating the beginning or middle of words such as ``\#\#'' for simple explanation.
}
\label{tbl:tokenization}
\end{table*}